\documentclass[a4paper,twoside]{article}

\newcommand{\hatI}[1]{#1\kern-0.5em\hat{\phantom{#1}}}

\usepackage{graphicx}

\usepackage{blindtext}
\usepackage{comment}
\usepackage{dsfont}
\usepackage{siunitx}
\usepackage[T1]{fontenc}
\usepackage[utf8]{inputenc}
\usepackage{subfigure}
\usepackage{multirow}
\usepackage{tabularx}
\usepackage{booktabs}
\usepackage{epsfig}
\usepackage{subcaption}
\usepackage{calc}
\usepackage{amssymb}
\usepackage{amstext}
\usepackage{amsmath}
\usepackage{amsthm}
\usepackage{multicol}
\usepackage{pslatex}
\usepackage{apalike}
\usepackage[ruled]{algorithm2e}
\usepackage[bottom]{footmisc}
\usepackage{url}
\usepackage{SCITEPRESS}     

\begin{document}

\title{Neural Network Meta Classifier:\\ Improving the Reliability of Anomaly Segmentation}

\author{\authorname{Jurica Runtas, Tomislav Petković}
\affiliation{University of Zagreb Faculty of Electrical Engineering and Computing, Unska 3, 10000 Zagreb, Croatia}
\email{\{jurica.runtas, tomislav.petkovic.jr\}@fer.unizg.hr}
}


\abstract{Deep neural networks (DNNs) are a contemporary solution for semantic segmentation and are usually trained to operate on a predefined closed set of classes. In open-set environments, it is possible to encounter semantically unknown objects or anomalies. Road driving is an example of such an environment in which, from a safety standpoint, it is important to ensure that a DNN indicates it is operating outside of its learned semantic domain. One possible approach to anomaly segmentation is entropy maximization, which is paired with a logistic regression based post-processing step called meta classification, which is in turn used to improve the reliability of detection of anomalous pixels. We propose to substitute the logistic regression meta classifier with a more expressive lightweight fully connected neural network. We analyze advantages and drawbacks of the proposed neural network meta classifier and demonstrate its better performance over logistic regression. We also introduce the concept of informative out-of-distribution examples which we show to improve training results when using entropy maximization in practice. Finally, we discuss the loss of interpretability and show that the behavior of logistic regression and neural network is strongly correlated. The code is publicly available at \url{https://github.com/JuricaRuntas/meta-ood}.}

\onecolumn \maketitle \normalsize \setcounter{footnote}{0} \vfill

\section{\uppercase{Introduction}}
\label{sec:introduction}

Semantic segmentation is a computer vision task in which each pixel of an image is assigned into one of predefined classes. An example of a real-world application is an autonomous driving system where semantic segmentation is an important component for visual perception of a driving environment \cite{DBLP:journals/corr/abs-2103-05445,DBLP:journals/corr/JanaiGBG17}.

Deep neural networks (DNNs) are a contemporary solution to the semantic segmentation task. DNNs are usually trained to operate on a predefined closed set of classes. However, this is in a contradiction with the nature of an environment in which aforementioned autonomous driving systems are deployed. Such systems operate in a so-called open-set environment where DNNs will encounter anomalies, i.e., objects that do not belong to any class from the predefined closed set of classes used during training \cite{DBLP:journals/corr/abs-1910-11296}.

From a safety standpoint, it is very important that a DNN classifies pixels of any encountered anomaly as anomalous and not as one of the predefined classes. The presence of an anomaly indicates that a DNN is operating outside of its learned semantic domain so a corresponding action may be taken, e.g., there is an unknown object on the road and an emergency braking procedure is initiated.

One approach to anomaly segmentation is entropy maximization  \cite{DBLP:journals/corr/abs-2012-06575}.
It is usually paired with a logistic regression based post-processing step called meta classification, which is used to improve the reliability of detection of anomalous pixels in the image, driving subsequent anomaly detection.

In this paper, we explore entropy maximization approach to anomaly segmentation where we propose to substitute the logistic regression meta classifier with a lightweight fully connected neural network.
Such a network is more expressive than the logistic regression meta classifier, so we expect an improvement in anomaly detection performance. Then, we provide additional analysis of the entropy maximization that shows that caution must be taken when using it in practice in order to ensure its effectiveness. To that end, we introduce the concept of informative out-of-distribution examples which we show to improve training results. Finally, we discuss the loss of interpretability and show that the behavior of logistic regression and neural network is strongly correlated, suggesting that the loss of interpretability may not be a significant drawback after all.

\section{\uppercase{Related Work}}

The task of identifying semantically anomalous regions in an image is called anomaly segmentation or, in the more general context, out-of-distribution (OoD) detection. Regardless of a specific method used for anomaly segmentation, the main objective is to obtain an anomaly segmentation score map. The anomaly segmentation score map $\mathbf{a}$ indicates the possibility of the presence of an anomaly at each pixel location where higher score indicates more probable anomaly \cite{chan2022detecting}. Methods described in the literature differ in the ways how such a map is obtained.

The methods described in the early works are based on the observation that anomalies usually result in low confidence predictions allowing for their detection. These methods include thresholding the maximum softmax probability \cite{DBLP:conf/iclr/HendrycksG17}, ODIN \cite{DBLP:conf/iclr/LiangLS18}, uncertainty estimation through the usage of Bayesian methods such as Monte-Carlo dropout \cite{gal2016dropout,DBLP:journals/corr/KendallBC15}, ensembles \cite{lakshminarayanan2017simple} and distance based uncertainty estimation through Mahalanobis distance \cite{DBLP:journals/corr/abs-1812-02765,lee2018simple} or Radial Basis Function Networks (RBFNs) \cite{DBLP:journals/corr/abs-2111-12866,DBLP:journals/corr/abs-2003-02037}. These methods do not rely on the utilization of negative datasets containing images with anomalies so they are classified as anomaly segmentation methods without outlier supervision. 

However, methods such as entropy maximization \cite{DBLP:journals/corr/abs-2012-06575} use entire images sampled from a negative dataset. Some methods cut and paste anomalies from images in the chosen negative dataset on the in-distribution images \cite{DBLP:journals/corr/abs-1908-01098,DBLP:journals/corr/abs-2101-09193,grcić2022densehybridhybridanomalydetection}. The negative images are used to allow the model to learn a representation of the unknown; therefore, such methods belong to the category of anomaly segmentation methods with outlier supervision.

Finally, there are methods that use generative models for the purpose of anomaly segmentation \cite{DBLP:journals/corr/abs-2103-05445,DBLP:journals/corr/abs-1904-03215,Grcic2021DenseAD,DBLP:journals/corr/abs-1904-07595,DBLP:journals/corr/abs-2003-08440}, usually through the means of reconstruction or normalizing flows, with or without outlier supervision. Current \hbox{state-of-the-art} anomaly segmentation methods \cite{ackermann2023maskomalyzeroshotmaskanomalysegmentation,rai2023unmaskinganomaliesroadscenesegmentation,nayal2023rbasegmentingunknownregions,delić2024outlierdetectionensemblinguncertainty} utilize mask-based semantic \mbox{segmentation} \cite{cheng2021perpixelclassificationneedsemantic,cheng2022maskedattentionmasktransformeruniversal}.

\section{\uppercase{Methodology}}
\label{sec:methodology}

In this section, we describe a method for anomaly segmentation called entropy maximization. Then, we describe a post-processing step called meta classification, which is used for improving the reliability of anomaly segmentation. Finally, we describe our proposed improvement to the original meta classification approach \cite{DBLP:journals/corr/abs-2012-06575}. All methods described in the following two subsections are introduced and thoroughly described in \cite{DBLP:journals/corr/abs-2012-06575,chan2022detecting,DBLP:journals/corr/abs-2005-06831,DBLP:journals/corr/abs-1811-00648,DBLP:journals/corr/abs-1904-04516}.

\subsection{Notation}

Let $\mathbf{x} \in [0, 1]^{H \times W \times3}$ denote a normalized color image of spatial dimensions $H \times W$. Let $\mathcal{I} = \{1, 2, ..., H\} \times \{1, 2, ..., W\}$ denote the set of pixel locations. Let $\mathcal{C} = \{1, 2, ..., C\}$ denote the set of $|\mathcal{C}|$ predefined classes. We define a set of training data used to train a semantic segmentation neural network in a supervised manner as $\mathcal{D}^{train}_{in} = \bigl\{(\mathbf{x}_{j}, \mathbf{m}_{j})\bigr\}^{N^{train}_{in}}_{j=1}$, where $N^{train}_{in}$ denotes the total number of in-distribution training samples and $\mathbf{m}_{j} = (m_{i})_{i \in \mathcal{I}} \in \mathcal{C}^{H \times W}$ is the corresponding ground truth segmentation mask of $\mathbf{x}_{j}$. Let $\mathbf{F}:~[0, 1]^{H \times W \times3} \rightarrow  [0, 1]^{H \times W \times |\mathcal{C}|}$ be a semantic segmentation neural network that produces pixel-wise class probabilities for a given image $\mathbf{x}$.

\subsection{Anomaly Segmentation via Entropy Maximization}

Let $\mathbf{p}_{i}(\mathbf{x}) = \bigl{(}p_{i}(c|\mathbf{x})\bigr{)}_{i \in \mathcal{I}, c \in \mathcal{C}}  \in [0, 1]^{|\mathcal{C}|}$ denote a vector of probabilities such that the $p_{i}(c|\mathbf{x})$ is a probability of a pixel location $i \in \mathcal{I}$ of a given image $\mathbf{x} \in \mathcal{D}_{in}$ being a pixel that belongs to the class $c \in \mathcal{C}$. We define $\mathbf{p}(\mathbf{x}) = \bigl{(}\mathbf{p}_{i}(\mathbf{x})\bigr{)}_{i \in \mathcal{I}} \in [0, 1]^{H \times W \times |\mathcal{C}|}$, the probability distribution over images in $\mathcal{D}_{in}$. When using $\mathcal{D}^{train}_{in}$ to train a semantic segmentation neural network \textbf{F}, one can interpret that the network is being trained to estimate $\mathbf{p}(\mathbf{x})$, denoted by $\hat{\mathbf{p}}(\mathbf{x})$. For a semantic segmentation network in the context of anomaly segmentation, it would be a desirable property if such network could output a high prediction uncertainty for OoD pixels which can in turn be quantified with a per-pixel entropy. For a given image $\mathbf{x} \in [0, 1]^{H \times W \times3}$ and a pixel location $i \in \mathcal{I}$, the per-pixel prediction entropy is defined as
\begin{equation}
	\label{eq:per_pixel_prediction_entropy}
	E_{i}\bigl(\mathbf{\hat{p}}_{i}(\mathbf{x})\bigr) = -\sum_{c \in \mathcal{C}}^{\phantom{c}}\hat p_{i}(c|\mathbf{x})\log{\bigl(\hat p_{i}(c|\mathbf{x})\bigr)}\text{,}
\end{equation}
where $E_{i}\bigl(\mathbf{\hat{p}}_{i}(\mathbf{x})\bigr)$ is maximized by the uniform (non-informative) probability distribution $\mathbf{\hat{p}}_{i}(\mathbf{x})$ which makes it an intuitive uncertainty measure.

We define a set of OoD training samples as $\mathcal{D}^{train}_{out} = \bigl\{(\mathbf{x}_{j}, \mathbf{m}_{j})\bigr\}^{N^{train}_{out}}_{j=1}$ where $N^{train}_{out}$ denotes the total number of such samples. In practice, $\mathcal{D}^{train}_{out}$ is a general-purpose dataset that contains diverse taxonomy exceeding the one found in the chosen domain-specific dataset $\mathcal{D}^{train}_{in}$ and it serves as a proxy for images containing anomalies.

It has been shown \cite{DBLP:journals/corr/abs-2012-06575} that one can make the output of a semantic segmentation neural network $\mathbf{F}$ have a high entropy on OoD pixel locations by employing a multi-criteria training objective defined as
\begin{equation}
    \label{eq:modified_training_objective}
    \begin{split}
    \mathcal{L} = (1 - \lambda)\cdot&\mathbb{E}_{(\mathbf{x}, \mathbf{m}) \in \mathcal{D}^{train}_{in}}\Bigl[l_{in}\bigl(\mathbf{F}(\mathbf{x}), \mathbf{m}\bigr)\Bigr] + \\  \lambda\cdot&\mathbb{E}_{(\mathbf{x}, \mathbf{m}) \in \mathcal{D}^{train}_{out}}\Bigl[l_{out}\bigl(\mathbf{F}(\mathbf{x}), \mathbf{m}\bigr)\Bigr]\text{,}
    \end{split}
\end{equation}
where $\lambda \in [0,1]$ is used for controlling the impact of each part of the overall objective.

When minimizing the overall objective defined by Eq.~(\ref{eq:modified_training_objective}), a commonly used cross-entropy loss is applied for in-distribution training samples defined as
\begin{equation}
    \label{eq:l_in_part_of_the_objective}
    l_{in}\bigl(\mathbf{F}(\mathbf{x}), \mathbf{m}\bigr) = - \sum_{i \in \mathcal{I}}^{\phantom{c}} \sum_{c \in \mathcal{C}} \mathds{1}_{m_{i}=c} \cdot \log{\bigl(\hat{p}_{i}(c|\mathbf{x})\bigr)}\text{,}
\end{equation}
where $\mathds{1}_{c=m_{i}} \in \{0,1\}$ is the indicator function being equal to one if the class index $c \in \mathcal{C}$ is, for a given pixel location $i \in \mathcal{I}$, equal to the class index $m_{i}$ defined by the ground truth segmentation mask \textbf{m} and zero otherwise. For OoD training samples, a slightly modified cross-entropy loss defined as
\begin{equation}
    \label{eq:l_out_part_of_the_objective}
    l_{out}\bigl(\mathbf{F}(\mathbf{x}), \mathbf{m}\bigr) = - \sum_{i \in \mathcal{I}_{out}}^{\phantom{c}} \sum_{c \in \mathcal{C}} \tfrac{1}{|\mathcal{C}|} \log{\bigl(\hat{p}_{i}(c|\mathbf{x})\bigr)}
\end{equation}
is applied for pixel locations $i \in \mathcal{I}$ labeled as OoD in the ground truth segmentation mask \textbf{m}. It can be shown \cite{DBLP:journals/corr/abs-2012-06575} that minimizing $l_{out}$ defined by Eq.~(\ref{eq:l_out_part_of_the_objective}) is equivalent to maximizing per-pixel prediction entropy $E_{i}\bigl(\mathbf{\hat{p}}_{i}(\mathbf{x})\bigr)$ defined by Eq.~(\ref{eq:per_pixel_prediction_entropy}), hence the name entropy maximization. The anomaly segmentation score map \textbf{a} can then be obtained by normalizing the per-pixel prediction entropy, i.e.,
\begin{equation}
    \label{eq:normalized_per_pixel_prediction_entropy}
    \mathbf{a} = (a_{i})_{i \in \mathcal{I}} \in [0, 1]^{H \times W},\: a_{i} = \frac{E_{i}\bigl(\mathbf{\hat{p}}_{i}(\mathbf{x})\bigr)}{\log{(|\mathcal{C}|)}}\text{.}
\end{equation}

\subsection{Meta Classification}

Meta classification is the task of discriminating between a false positive prediction and a true positive prediction. Training a network with a modified entropy maximization training objective increases the network's sensitivity towards predicting OoD objects and can result in a substantial number of false positive predictions \cite{chan2019metafusioncontrolledfalsenegativereduction,DBLP:journals/corr/abs-2012-06575}. Applying meta classification in order to post-process the network's prediction has been shown to significantly improve the network's ability to reliably detect OoD objects. For a given image $\mathbf{x}$, we define a set of pixel locations being predicted as OoD as 
\begin{equation}
    \label{eq:a_set_of_OoD_pixel_locations}
    \hatI{\mathcal{I}}_{out}(\mathbf{x}, \mathbf{a}) = \bigl\{i \in \mathcal{I} \mid a_{i} \ge t, t \in [0,1]\bigr\}
\end{equation}
where $t$ represents a fixed threshold and \textbf{a} is computed using Eq.~(\ref{eq:normalized_per_pixel_prediction_entropy}). Based on $\hatI{\mathcal{I}}_{out}(\mathbf{x}, \mathbf{a})$, a set of connected components representing OoD object predictions defined as $\hat{\mathcal{K}}(\mathbf{x}, \mathbf{a}) \subseteq \mathcal{P}\bigl(\hatI{\mathcal{I}}_{out}(\mathbf{x}, \mathbf{a})\bigr)$ is constructed. Note that $\mathcal{P}\bigl(\hatI{\mathcal{I}}_{out}(\mathbf{x}, \mathbf{a})\bigr)$ denotes the power set of $\hatI{\mathcal{I}}_{out}(\mathbf{x}, \mathbf{a})$.

Meta classifier is a lightweight model added on top of a semantic segmentation network \textbf{F}. After training \textbf{F} for entropy maximization on the pixels of known OoD objects, a structured dataset of hand-crafted metrics is constructed. For every OoD object prediction $\hat{k} \in \hat{\mathcal{K}}(\mathbf{x}, \mathbf{a})$, different pixel-wise uncertainty measures are derived solely from $\mathbf{\hat{p}}(\mathbf{x})$ such as normalized per-pixel prediction entropy of Eq.~(\ref{eq:per_pixel_prediction_entropy}), maximum softmax probability, etc. In addition to metrics derived from $\mathbf{\hat{p}}(\mathbf{x})$, metrics based on the OoD object prediction geometry features are also included such as the number of pixels contained in $\hat{k}$, various ratios regarding interior and boundary pixels, geometric center, geometric features regarding the neighborhood of $\hat{k}$, etc. \cite{DBLP:journals/corr/abs-2012-06575,DBLP:journals/corr/abs-1811-00648}. 

After a dataset with the hand-crafted metrics is constructed, a meta classifier is trained to classify OoD object predictions in one of the following two sets,
\begin{equation}
    \label{eq:meta_classification_sets}
    \begin{split}
    &C_{TP}(\mathbf{x}, \mathbf{a}) = \bigl\{\hat{k} \in \hat{\mathcal{K}}(\mathbf{x}, \mathbf{a}) \mid IoU(\hat{k}, \mathbf{m}) > 0\bigr\}\text{ and} \\	
    &C_{FP}(\mathbf{x}, \mathbf{a}) = \bigl\{\hat{k} \in \hat{\mathcal{K}}(\mathbf{x}, \mathbf{a}) \mid IoU(\hat{k}, \mathbf{m}) = 0\bigr\}\text{,}
    \end{split}
\end{equation}
where $C_{TP}$ represents a set of true positive OoD object predictions, $C_{FP}$ a set of false positive OoD object predictions and $IoU$ represents the intersection over union of a OoD object prediction $\hat{k}$ with the corresponding ground truth segmentation mask $\mathbf{m}$. Each $(\mathbf{x}, \mathbf{m}) \in \mathcal{D}^{meta}_{out}$ is an element of a dataset containing known OoD objects used to train a meta classifier.

During inference, a meta classifier predicts whether an OoD object predictions obtained from \textbf{F} are false positive. Certainly, the prediction is done without the access to the ground truth segmentation mask \textbf{m} and is based on learned statistical and geometrical properties of the OoD object predictions obtained from the known unknowns. OoD object predictions classified as false positive are then removed and the final prediction is obtained.

\subsection{Neural Network Meta Classifier}
\label{subsec:neural_network_meta_classifier}

In \cite{DBLP:journals/corr/abs-2012-06575}, authors use logistic regression for the purpose of meta classifying predicted OoD objects. Their main argument for the use of logistic regression is that since it is a linear model, it is possible to analyze the impact of each hand-crafted metric used as an input to the model with an algorithm such as Least Angle Regression (LARS) \cite{Efron_2004}. However, we argue that even though it is desirable to have an interpretable model in order to analyze the relevance and the impact of its input, it is possible to achieve a significantly greater performance by employing a more expressive type of model such as a neural network.

Let $\hat{K}$ be a set containing OoD object predictions for every $(\mathbf{x}, \mathbf{m}) \in \mathcal{D}^{meta}_{out}$ defined as $\hat{K} =~\bigcup_{(\mathbf{x}, \mathbf{m}) \in \mathcal{D}^{meta}_{out}}\hat{\mathcal{K}}(\mathbf{x}, \mathbf{a})$. We formally define the aforementioned hand-crafted metrics dataset as $\mu \subset~\mathbb{R}^{|\hat{K}| \times N_{m}}$, where $N_{m}$ is the total number of hand-crafted metrics derived from each OoD object prediction.

We propose that instead of logistic regression as a meta classifier, a lightweight fully connected neural network is employed. Let $\mathbf{F}^{meta} : \mu \rightarrow [0, 1]$ denote such a neural network. We can interpret that $\mathbf{F}^{meta}$ outputs the probability of a given OoD object prediction being false positive based on the corresponding derived hand-crafted metrics according to Eq.~(\ref{eq:meta_classification_sets}). Let $p^{F}$ denote such probability. Since $\mathbf{F}^{meta}$ is essentially a binary classifier, we can train it using the binary cross-entropy loss defined as
\begin{equation}
	\label{eq:binary_cross_entropy_loss_meta_classifier}
	\mathcal{L}^{meta} = -\sum^{N}_{i = 1}y_{i}\log{\bigl(p^{F}_{i}\bigr)} + (1-y_{i})\log{\bigl(1 - p^{F}_{i}\bigr)}\text{,}
\end{equation}
where $N$ represents the number of OoD object predictions included in a mini-batch and $y_{i}$ represents the ground truth label of a given OoD object prediction and is equal to one if given OoD object prediction is false positive and zero otherwise.

\section{\uppercase{Experiments}}
\label{sec:experiments}

In this section, we briefly describe the experimental setup and evaluate our proposed neural network meta~classifier. 

\subsection{Experimental Setup}

For the purpose of the entropy maximization, we use DeepLabv3+ semantic segmentation model \cite{DBLP:journals/corr/abs-1802-02611} with a WideResNet38 backbone \cite{DBLP:journals/corr/WuSH16e} trained by Nvidia \cite{DBLP:journals/corr/abs-1812-01593}. The model is pretrained on Cityscapes dataset \cite{DBLP:journals/corr/CordtsORREBFRS16}. The pretrained model is fine-tuned according to Eq.~(\ref{eq:modified_training_objective}). We use Cityscapes dataset \cite{DBLP:journals/corr/CordtsORREBFRS16} as $\mathcal{D}^{train}_{in}$ containing 2,975 images while for $\mathcal{D}^{train}_{out}$ we use a subset of COCO dataset \cite{DBLP:journals/corr/LinMBHPRDZ14} which we denote as COCO-OoD. For the purpose of $\mathcal{D}^{train}_{out}$, we exclude images containing class instances that are also found in Cityscapes dataset. After filtering, 46,751 images remain. The model is trained for 4 epochs on random square crops of height and width of 480 pixels. Images that have height or width smaller than 480 pixels are resized. Before each epoch, we randomly shuffle 2,975 images from Cityscapes dataset with 297 images randomly sampled from the remaining 46,751 COCO images. Hyperparameters are set according to the baseline \cite{DBLP:journals/corr/abs-2012-06575}: loss weight $\lambda = 0.9$, entropy threshold $t = 0.7$. Adam optimizer \cite{kingma2017adam} is used with learning rate $\eta = \num{1e-5}$.

\subsection{Evaluation of Neural Network Meta Classiﬁer}
\begin{figure*}[htb]
	\centering
	\includegraphics[width=1\linewidth,height=0.4\linewidth]{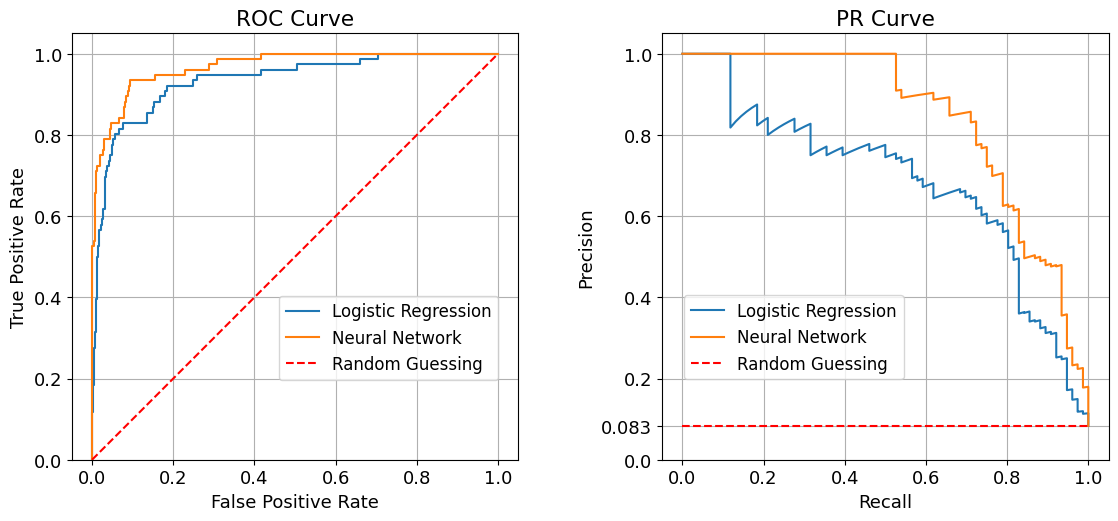} 
	\caption{ROC and PR meta classifier curves for OoD object predictions of LostAndFound Test images. On the PR curve, random guessing is represented as a constant dashed red line whose value is equal to the ratio of the number of OoD objects and the total number of predicted OoD objects.}
	\label{fig:roc_and_pr_meta_classifier_curves}
\end{figure*}

We use \cite{DBLP:journals/corr/abs-2012-06575} as a baseline. We substitute the logistic regression with a lightweight fully connected neural network whose architecture is shown in Table \ref{table:neural_network_meta_classifier_architecture}. The proposed meta classifier is trained on the hand-crafted metrics derived from OoD object predictions of the images in LostAndFound Test \cite{DBLP:journals/corr/PinggeraRGFRM16}. Derived hand-crafted metrics, i.e., corresponding OoD object predictions are leave-one-out cross validated according to Eq.~(\ref{eq:meta_classification_sets}). The meta classifier is trained using Adam optimizer with learning rate $\eta = \num{1e-3}$ and weight decay $\gamma = \num{5e-3}$ for 50 epochs with a mini-batch size $N = 128$. Note that in our case, the total number of hand-crafted metrics~$N_{m} = 75$. Also note that the logistic regression meta classifier has 76 parameters. The results are shown in Table~\ref{table:meta_classifier_comparison} and Fig.~\ref{fig:roc_and_pr_meta_classifier_curves}. In our experiments, the improved performance is especially noticeable when considering OoD object predictions consisting of a very small number of pixels.

\begin{table}[htb]
	\centering
    \fontsize{9pt}{9pt}\selectfont
    \begin{tabular}{c c c} 
        \toprule
        Layer & \# of neurons & \# of parameters \\
        \midrule
        \midrule
        Input layer & 75 & 5,700 \\
        \midrule
        1. layer & 75 & 5,700 \\
        \midrule
        2. layer & 75 & 5,700 \\
        \midrule
        Output layer & 1 & 76 \\
    \bottomrule
    \end{tabular}
	\caption{Architecture of the neural network meta classifier. All layers are fully connected and a sigmoid activation is applied after the last layer. The total number of parameters is 17,176.}
	\label{table:neural_network_meta_classifier_architecture}
\end{table}

\begin{table}[htb]
	\centering
	\resizebox{\linewidth}{!}{
        \fontsize{9pt}{9pt}\selectfont
		\begin{tabular}{c c c c} 
			\toprule
			Model Type & \multicolumn{2}{c}{Logistic Regression} & Neural Network \\
			\midrule
			\midrule
			Source & Baseline & Reproduced & Ours \\
			\midrule
			AUROC & 0.9444 & 0.9342 & \textbf{0.9680} \\
			AUPRC & 0.7185 & 0.6819 & \textbf{0.8418} \\
			\bottomrule
		\end{tabular}
	}
	\caption{Performance comparison of meta classifiers. Note that the given results are based on OoD object predictions obtained with entropy threshold $t = 0.7$ of Eq.~(\ref{eq:a_set_of_OoD_pixel_locations}).}
	\label{table:meta_classifier_comparison}
\end{table}

\section{\uppercase{Discussion}}
\label{sec:discussion}

In this section, we introduce the notion of high and low informative OoD proxy images, and we show that the high informative proxy OoD images are the ones from which the semantic segmentation network can learn to reliably output high entropy on OoD pixels of images seen during inference. Then, we discuss the loss of interpretability, a drawback of using the proposed neural network meta classifier instead of the interpretable logistic regression meta classifier and show that it may not be a significant drawback after~all.

\subsection{On Outlier Supervision of the Entropy Maximization}

\begin{figure*}[htb]
	\centering
	\subfigure[Examples of high informative proxy OoD images.]{\includegraphics[width=0.49\linewidth, height=0.25\linewidth]{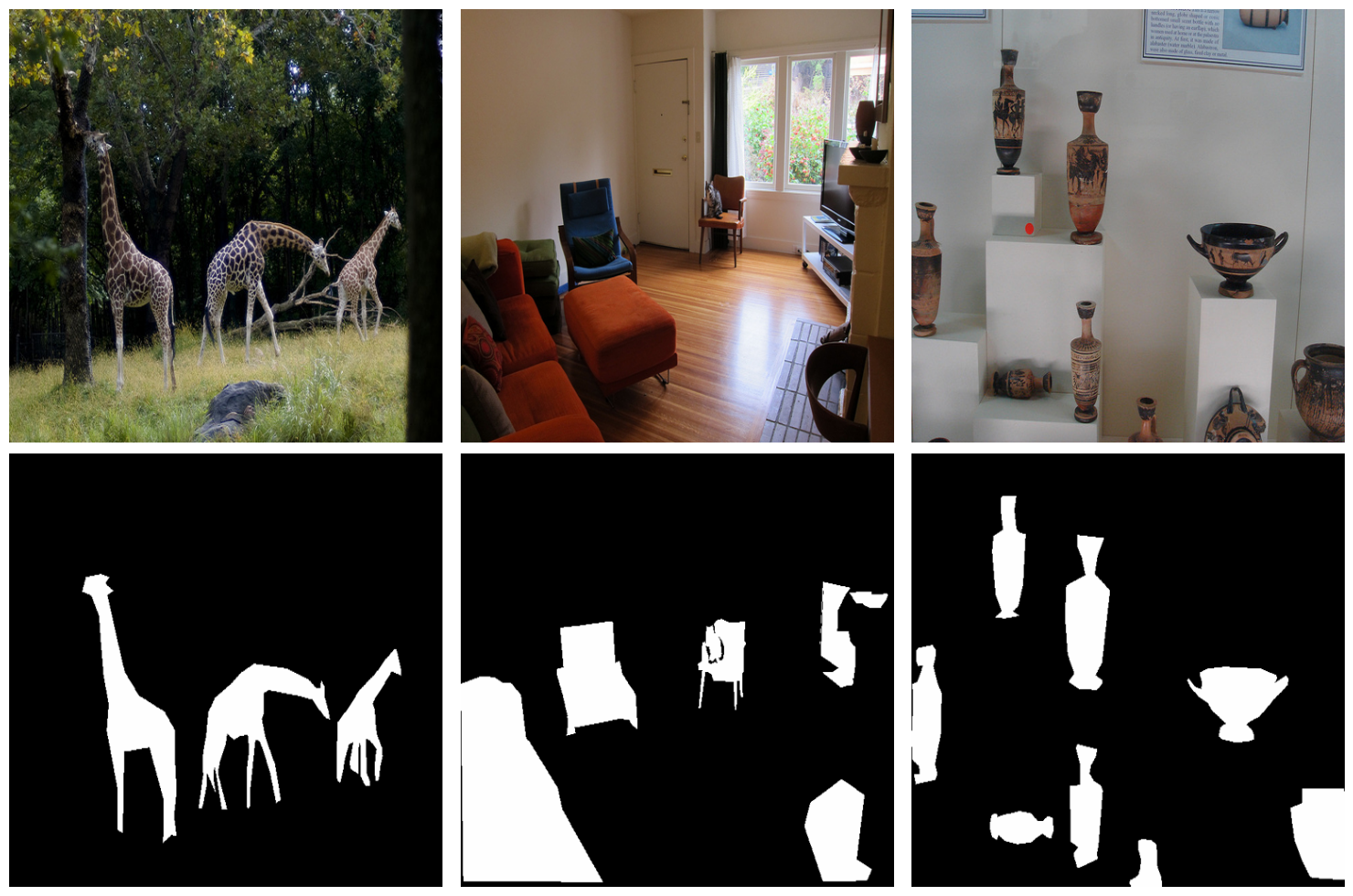}}
	\hspace*{\fill}
	\subfigure[Examples of low informative proxy OoD images.]{\includegraphics[width=0.49\linewidth, height=0.25\linewidth]{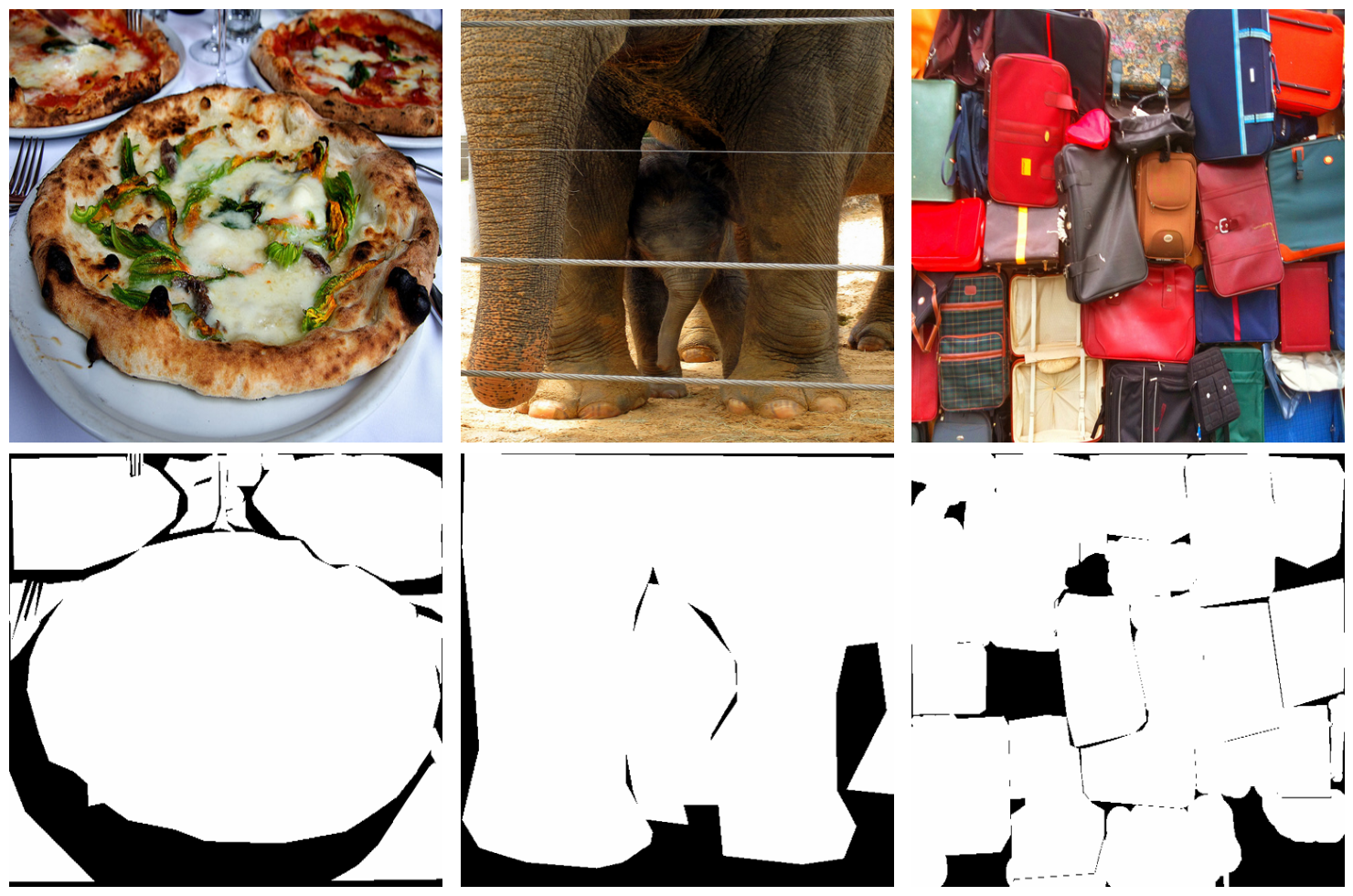}}
	\caption{Examples of high and low informative proxy OoD images. The first row contains the proxy OoD images while the second row contains ground truth segmentation masks such that the white regions represent pixels labeled as OoD for which Eq.~(\ref{eq:l_out_part_of_the_objective}) is applied.}
	\label{fig:high_and_low_informative_proxy_OoD_images}
\end{figure*}

We introduce the notion of high informative and low informative proxy OoD images. What we mean by high and low informative is illustrated with Fig.~\ref{fig:high_and_low_informative_proxy_OoD_images}. We have noticed empirically that high informative proxy OoD images have two important characteristics that differentiate them from the low informative proxy OoD images: spatially clear separation between objects and clear object boundaries. 

Our conjecture is that the low informative proxy OoD images have little to no impact on the entropy maximization training or can even negatively impact the training procedure. On the other hand, high informative proxy OoD images are the ones from which the semantic segmentation network can learn to reliably output high entropy on OoD pixels of images seen during inference, denoted by $\mathcal{D}_{out} \setminus \mathcal{D}^{train}_{out}$, where~$\setminus$ represents the set difference.

\begin{table*}[htb]
	\centering
	\resizebox{\linewidth}{!}{
		\begin{tabular}{c c c c c c c c c} 
			\toprule
			Metric & \multicolumn{4}{c}{$\text{FPR}_{95}$} & \multicolumn{4}{c}{AUPRC} \\ 
			\midrule
			\midrule
			Source & DLV3+W38 & COCO-OoD & L-20\%-OoD & M-80\%-OoD & DLV3+W38 & COCO-OoD & L-20\%-OoD & M-80\%-OoD \\
			\midrule
			LostAndFound Test & 0.35 & 0.15 & \textbf{0.09} & 0.13 & 0.46 & 0.75 & \textbf{0.78} & 0.48 \\
			Fishyscapes Static & 0.19 & 0.17 & \textbf{0.12} & 0.31 & 0.25 & 0.64 & \textbf{0.73} & 0.25 \\
			\bottomrule
		\end{tabular}
	}
	\caption{Results for the entropy maximization training using COCO-OoD subsets. Column DLV3+W38 contains the results obtained from the model used for fine-tuning \protect{\cite{DBLP:journals/corr/abs-1812-01593}} which was trained exclusively on the in-distribution images. Other columns contain results obtained from the best model after performing the entropy maximization training numerous times with a given subset.}
	\label{table:coco_subsets_training_results}
\end{table*}

To investigate our conjecture, we perform the entropy maximization training on subsets of COCO-OoD. We consider it difficult to universally quantify the mentioned characteristics of high informative proxy OoD images, however, we notice a significant correlation between the percentage of the labeled OoD pixels and the desirable properties found in high informative OoD proxy images. We use COCO-OoD proxy for the creation of the two disjoint sets such that the first contains images from COCO-OoD that have at most 20\% of pixels labeled as OoD (denoted as L-20\%-OoD) and the second that contains images from COCO-OoD that have at least 80\% of pixels labeled as OoD (denoted as M-80\%-OoD). Table~\ref{table:coco_subsets_training_results} shows that performing the entropy maximization training using M-80\%-OoD results in a little to no improvement in comparison to the model trained exclusively on the in-distribution images. On the other hand, using L-20\%-OoD produced even better results than the ones obtained with the usage of COCO-OoD.

\subsection{Interpretability of Neural Network Meta Classifier}

\begin{figure}[htb]
    \centering
    \includegraphics[width=1\linewidth]{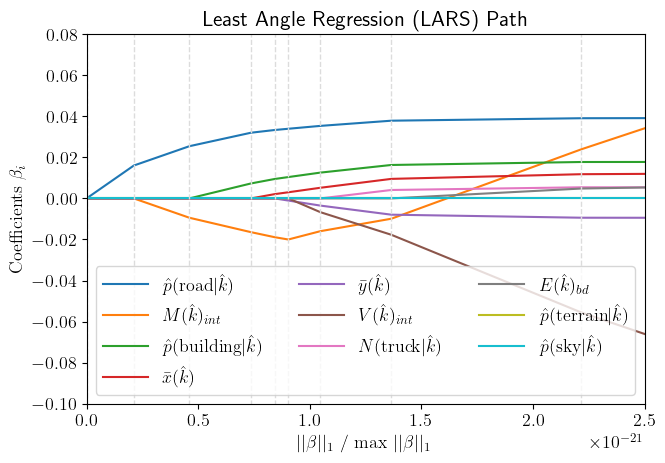}
    \caption{LARS path for the hand-crafted metrics at $t = 0.7$. A detailed description of the hand-crafted metrics can be found in \protect{\cite{DBLP:journals/corr/abs-2012-06575}}.}
    \label{fig:lars}
\end{figure}

\begin{figure*}[htb]
	\centering
	\includegraphics[width=1\linewidth]{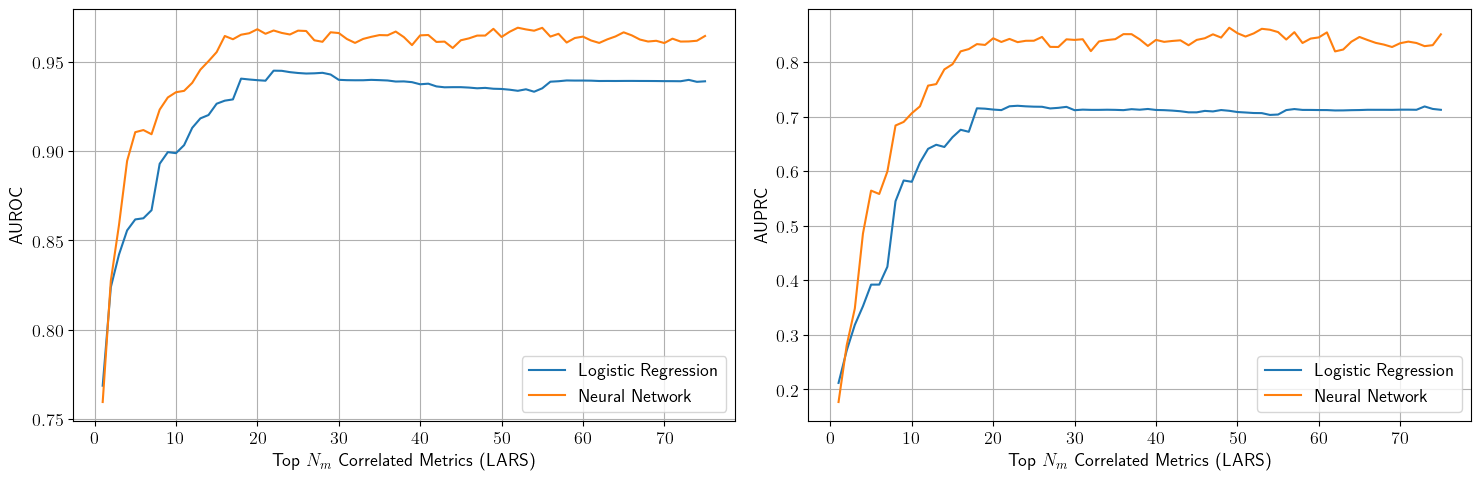}
	\caption{Performance comparison of logistic regression meta classifier and neural network meta classifier when trained on subsets of the hand-crafted metrics dataset $\mu$. For each value $N_{m}$ on the x-axis, we train the meta classifiers on the subset of $\mu$ such that we take the first $N_{m}$ metrics having the most correlation with the response according to LARS.}
	\label{fig:most_correlated_auprc_auroc_graph}
\end{figure*}

A drawback of using a neural network as a meta classifier is the loss of interpretability. However, we attempt to further understand the performance of our proposed meta classifier. Fig.~\ref{fig:lars} shows LARS path for ten hand-crafted metrics most correlated with the response of logistic regression, i.e., the ones which contribute the most in classifying OoD object predictions. One can interpret LARS as a way of sorting the hand-crafted metrics based on the impact on the response of logistic regression. Algorithm \ref{alg:alg1} offers a way to leverage this kind of reasoning in order to gain a further insight in how neural network meta classifier behaves in comparison to logistic regression. Note that we assume that LARS sorts hand-crafted metrics in descending order with respect to the correlation. Fig.~\ref{fig:most_correlated_auprc_auroc_graph} shows results of executing Algorithm \ref{alg:alg1} for both meta classifiers. 

For the logistic regression meta classifier, after we take a subset of $\mu$ containing 21 most correlated hand-crafted metrics according to LARS, adding \mbox{remaining} hand-crafted metrics results in little to no improvement in performance. We can see that the neural network meta classifier exhibits a similar behavior, although in a more unstable manner. The \mbox{obvious} difference in performance can be most likely attributed to the fact that neural network meta classifier is more expressive and better aggregates the hand-crafted metrics. We argue that the hand-crafted metrics having the most impact on the performance of logistic regression meta classifier also do so in the case of neural network meta classifier. Such insight could alleviate presumably the most significant drawback of using neural network meta classifier instead of logistic regression meta classifier - the loss of \mbox{interpretability}.

\begin{algorithm}[htb]
 \SetKwInOut{Input}{Input}
 \SetKwInOut{Output}{Output}
 \caption{Incremental meta classifier evaluation.}
 \label{alg:alg1}
 \Input{$\mathbf{F}^{meta}$, $\mu$, $N_{m}$}
 \Output{lists of AUROC and AUPRC metrics}
 AUROC $\longleftarrow$ $[\:]$\;
 AUPRC $\longleftarrow$ $[\:]$\;
 MetricsSortedByCorrelation $\longleftarrow$ LARS($\mu$)\;
 \For{$i=1$ \KwTo $N_{m}$}
 {  
    $\xi \longleftarrow$ MetricsSortedByCorrelation[:i]\;
    initializeModel($\mathbf{F}^{meta}$)\;
    trainModel($\mathbf{F}^{meta}$, $\xi$)\;
    ($m_{1}$, $m_{2}$) $\longleftarrow$ evaluateModel($\mathbf{F}^{meta}$, $\xi$)\;
    AUROC.append($m_{1}$)\;
    AUPRC.append($m_{2}$)\;
 }
\end{algorithm}

\section{\uppercase{Conclusion}}
\label{sec:conclusion}

In this paper, we explored the anomaly segmentation method called entropy maximization which can increase the network's sensitivity towards predicting OoD objects, but which can also result in a substantial number of false positive predictions. Hence, the meta classification post-processing step is applied in order to improve the network's ability to reliably detect OoD objects. Our experimental results showed that employing the proposed neural network meta classifier results in a significantly greater performance in comparison to the logistic regression meta classifier.

Furthermore, we provided additional analysis of the entropy maximization training which showed that in order to ensure its effectiveness, caution must be taken when choosing which images are going to be used as proxy OoD images. Our experimental results demonstrated that high informative proxy OoD images are the ones from which the \mbox{semantic segmentation} network can learn to reliably output high entropy on OoD pixels of images seen during inference and are therefore more beneficial to the entropy maximization training in terms of how well a semantic segmentation neural network can detect OoD objects afterwards.

Finally, a drawback of using the neural network meta classifier is the loss of interpretability. In our attempt to further analyze the performance of the proposed neural network meta classifier, we found that the behavior of logistic regression and neural network is strongly correlated, suggesting that the loss of interpretability may not be a significant drawback after~all.

\bibliographystyle{apalike}
{\small
\bibliography{references}}

\begin{thebibliography}{}

\bibitem[Ackermann et~al., 2023]{ackermann2023maskomalyzeroshotmaskanomalysegmentation}
Ackermann, J., Sakaridis, C., and Yu, F. (2023).
\newblock Maskomaly:zero-shot mask anomaly segmentation.

\bibitem[Bevandić et~al., 2019]{DBLP:journals/corr/abs-1908-01098}
Bevandić, P., Krešo, I., Oršić, M., and Šegvić, S. (2019).
\newblock Simultaneous semantic segmentation and outlier detection in presence of domain shift.
\newblock {\em CoRR}, abs/1908.01098.

\bibitem[Bevandić et~al., 2021]{DBLP:journals/corr/abs-2101-09193}
Bevandić, P., Krešo, I., Oršić, M., and Šegvić, S. (2021).
\newblock Dense outlier detection and open-set recognition based on training with noisy negative images.
\newblock {\em CoRR}, abs/2101.09193.

\bibitem[Biase et~al., 2021]{DBLP:journals/corr/abs-2103-05445}
Biase, G.~D., Blum, H., Siegwart, R., and Cadena, C. (2021).
\newblock Pixel-wise anomaly detection in complex driving scenes.
\newblock {\em CoRR}, abs/2103.05445.

\bibitem[Blum et~al., 2019]{DBLP:journals/corr/abs-1904-03215}
Blum, H., Sarlin, P., Nieto, J.~I., Siegwart, R., and Cadena, C. (2019).
\newblock The fishyscapes benchmark: Measuring blind spots in semantic segmentation.
\newblock {\em CoRR}, abs/1904.03215.

\bibitem[Chan et~al., 2020]{DBLP:journals/corr/abs-2012-06575}
Chan, R., Rottmann, M., and Gottschalk, H. (2020).
\newblock Entropy maximization and meta classification for out-of-distribution detection in semantic segmentation.
\newblock {\em CoRR}, abs/2012.06575.

\bibitem[Chan et~al., 2019]{chan2019metafusioncontrolledfalsenegativereduction}
Chan, R., Rottmann, M., Hüger, F., Schlicht, P., and Gottschalk, H. (2019).
\newblock Metafusion: Controlled false-negative reduction of minority classes in semantic segmentation.

\bibitem[Chan et~al., 2022]{chan2022detecting}
Chan, R., Uhlemeyer, S., Rottmann, M., and Gottschalk, H. (2022).
\newblock Detecting and learning the unknown in semantic segmentation.

\bibitem[Chen et~al., 2018]{DBLP:journals/corr/abs-1802-02611}
Chen, L., Zhu, Y., Papandreou, G., Schroff, F., and Adam, H. (2018).
\newblock Encoder-decoder with atrous separable convolution for semantic image segmentation.
\newblock {\em CoRR}, abs/1802.02611.

\bibitem[Cheng et~al., 2022]{cheng2022maskedattentionmasktransformeruniversal}
Cheng, B., Misra, I., Schwing, A.~G., Kirillov, A., and Girdhar, R. (2022).
\newblock Masked-attention mask transformer for universal image segmentation.

\bibitem[Cheng et~al., 2021]{cheng2021perpixelclassificationneedsemantic}
Cheng, B., Schwing, A.~G., and Kirillov, A. (2021).
\newblock Per-pixel classification is not all you need for semantic segmentation.

\bibitem[Cordts et~al., 2016]{DBLP:journals/corr/CordtsORREBFRS16}
Cordts, M., Omran, M., Ramos, S., Rehfeld, T., Enzweiler, M., Benenson, R., Franke, U., Roth, S., and Schiele, B. (2016).
\newblock The cityscapes dataset for semantic urban scene understanding.
\newblock {\em CoRR}, abs/1604.01685.

\bibitem[Delić et~al., 2024]{delić2024outlierdetectionensemblinguncertainty}
Delić, A., Grcić, M., and Šegvić, S. (2024).
\newblock Outlier detection by ensembling uncertainty with negative objectness.

\bibitem[Denouden et~al., 2018]{DBLP:journals/corr/abs-1812-02765}
Denouden, T., Salay, R., Czarnecki, K., Abdelzad, V., Phan, B., and Vernekar, S. (2018).
\newblock Improving reconstruction autoencoder out-of-distribution detection with mahalanobis distance.
\newblock {\em CoRR}, abs/1812.02765.

\bibitem[Efron et~al., 2004]{Efron_2004}
Efron, B., Hastie, T., Johnstone, I., and Tibshirani, R. (2004).
\newblock Least angle regression.
\newblock {\em The Annals of Statistics}, 32(2).

\bibitem[Gal and Ghahramani, 2016]{gal2016dropout}
Gal, Y. and Ghahramani, Z. (2016).
\newblock Dropout as a bayesian approximation: Representing model uncertainty in deep learning.

\bibitem[Grcić et~al., 2021]{Grcic2021DenseAD}
Grcić, M., Bevandić, P., and Šegvić, S. (2021).
\newblock Dense anomaly detection by robust learning on synthetic negative data.
\newblock {\em ArXiv}, abs/2112.12833.

\bibitem[Grcić et~al., 2022]{grcić2022densehybridhybridanomalydetection}
Grcić, M., Bevandić, P., and Šegvić, S. (2022).
\newblock Densehybrid: Hybrid anomaly detection for dense open-set recognition.

\bibitem[Hendrycks and Gimpel, 2017]{DBLP:conf/iclr/HendrycksG17}
Hendrycks, D. and Gimpel, K. (2017).
\newblock A baseline for detecting misclassified and out-of-distribution examples in neural networks.
\newblock In {\em 5th International Conference on Learning Representations, {ICLR} 2017, Toulon, France, April 24-26, 2017, Conference Track Proceedings}. OpenReview.net.

\bibitem[Janai et~al., 2020]{DBLP:journals/corr/JanaiGBG17}
Janai, J., G\"{u}ney, F., Behl, A., and Geiger, A. (2020).
\newblock Computer vision for autonomous vehicles: Problems, datasets and state of the art.
\newblock {\em Found. Trends. Comput. Graph. Vis.}, 12(1–3):1–308.

\bibitem[Kendall et~al., 2015]{DBLP:journals/corr/KendallBC15}
Kendall, A., Badrinarayanan, V., and Cipolla, R. (2015).
\newblock Bayesian segnet: Model uncertainty in deep convolutional encoder-decoder architectures for scene understanding.
\newblock {\em CoRR}, abs/1511.02680.

\bibitem[Kingma and Ba, 2017]{kingma2017adam}
Kingma, D.~P. and Ba, J. (2017).
\newblock Adam: A method for stochastic optimization.

\bibitem[Lakshminarayanan et~al., 2017]{lakshminarayanan2017simple}
Lakshminarayanan, B., Pritzel, A., and Blundell, C. (2017).
\newblock Simple and scalable predictive uncertainty estimation using deep ensembles.

\bibitem[Lee et~al., 2018]{lee2018simple}
Lee, K., Lee, K., Lee, H., and Shin, J. (2018).
\newblock A simple unified framework for detecting out-of-distribution samples and adversarial attacks.

\bibitem[Li and Kosecka, 2021]{DBLP:journals/corr/abs-2111-12866}
Li, Y. and Kosecka, J. (2021).
\newblock Uncertainty aware proposal segmentation for unknown object detection.
\newblock {\em CoRR}, abs/2111.12866.

\bibitem[Liang et~al., 2018]{DBLP:conf/iclr/LiangLS18}
Liang, S., Li, Y., and Srikant, R. (2018).
\newblock Enhancing the reliability of out-of-distribution image detection in neural networks.
\newblock In {\em 6th International Conference on Learning Representations, {ICLR} 2018, Vancouver, BC, Canada, April 30 - May 3, 2018, Conference Track Proceedings}. OpenReview.net.

\bibitem[Lin et~al., 2014]{DBLP:journals/corr/LinMBHPRDZ14}
Lin, T., Maire, M., Belongie, S.~J., Bourdev, L.~D., Girshick, R.~B., Hays, J., Perona, P., Ramanan, D., Doll{\'{a}}r, P., and Zitnick, C.~L. (2014).
\newblock Microsoft {COCO:} common objects in context.
\newblock {\em CoRR}, abs/1405.0312.

\bibitem[Lis et~al., 2019]{DBLP:journals/corr/abs-1904-07595}
Lis, K., Nakka, K.~K., Fua, P., and Salzmann, M. (2019).
\newblock Detecting the unexpected via image resynthesis.
\newblock {\em CoRR}, abs/1904.07595.

\bibitem[Nayal et~al., 2023]{nayal2023rbasegmentingunknownregions}
Nayal, N., Yavuz, M., Henriques, J.~F., and Güney, F. (2023).
\newblock Rba: Segmenting unknown regions rejected by all.

\bibitem[Oberdiek et~al., 2020]{DBLP:journals/corr/abs-2005-06831}
Oberdiek, P., Rottmann, M., and Fink, G.~A. (2020).
\newblock Detection and retrieval of out-of-distribution objects in semantic segmentation.
\newblock {\em CoRR}, abs/2005.06831.

\bibitem[Pinggera et~al., 2016]{DBLP:journals/corr/PinggeraRGFRM16}
Pinggera, P., Ramos, S., Gehrig, S., Franke, U., Rother, C., and Mester, R. (2016).
\newblock Lost and found: Detecting small road hazards for self-driving vehicles.
\newblock {\em CoRR}, abs/1609.04653.

\bibitem[Rai et~al., 2023]{rai2023unmaskinganomaliesroadscenesegmentation}
Rai, S.~N., Cermelli, F., Fontanel, D., Masone, C., and Caputo, B. (2023).
\newblock Unmasking anomalies in road-scene segmentation.

\bibitem[Rottmann et~al., 2018]{DBLP:journals/corr/abs-1811-00648}
Rottmann, M., Colling, P., Hack, T., H{\"{u}}ger, F., Schlicht, P., and Gottschalk, H. (2018).
\newblock Prediction error meta classification in semantic segmentation: Detection via aggregated dispersion measures of softmax probabilities.
\newblock {\em CoRR}, abs/1811.00648.

\bibitem[Rottmann and Schubert, 2019]{DBLP:journals/corr/abs-1904-04516}
Rottmann, M. and Schubert, M. (2019).
\newblock Uncertainty measures and prediction quality rating for the semantic segmentation of nested multi resolution street scene images.
\newblock {\em CoRR}, abs/1904.04516.

\bibitem[van Amersfoort et~al., 2020]{DBLP:journals/corr/abs-2003-02037}
van Amersfoort, J., Smith, L., Teh, Y.~W., and Gal, Y. (2020).
\newblock Simple and scalable epistemic uncertainty estimation using a single deep deterministic neural network.
\newblock {\em CoRR}, abs/2003.02037.

\bibitem[Wong et~al., 2019]{DBLP:journals/corr/abs-1910-11296}
Wong, K., Wang, S., Ren, M., Liang, M., and Urtasun, R. (2019).
\newblock Identifying unknown instances for autonomous driving.
\newblock {\em CoRR}, abs/1910.11296.

\bibitem[Wu et~al., 2016]{DBLP:journals/corr/WuSH16e}
Wu, Z., Shen, C., and van~den Hengel, A. (2016).
\newblock Wider or deeper: Revisiting the resnet model for visual recognition.
\newblock {\em CoRR}, abs/1611.10080.

\bibitem[Xia et~al., 2020]{DBLP:journals/corr/abs-2003-08440}
Xia, Y., Zhang, Y., Liu, F., Shen, W., and Yuille, A.~L. (2020).
\newblock Synthesize then compare: Detecting failures and anomalies for semantic segmentation.
\newblock {\em CoRR}, abs/2003.08440.

\bibitem[Zhu et~al., 2018]{DBLP:journals/corr/abs-1812-01593}
Zhu, Y., Sapra, K., Reda, F.~A., Shih, K.~J., Newsam, S.~D., Tao, A., and Catanzaro, B. (2018).
\newblock Improving semantic segmentation via video propagation and label relaxation.
\newblock {\em CoRR}, abs/1812.01593.

\end{thebibliography}
\end{document}